\documentclass[11pt]{article}

\usepackage[preprint]{acl}

\usepackage{times}
\usepackage{latexsym}

\usepackage[T1]{fontenc}

\usepackage[utf8]{inputenc}

\usepackage{microtype}

\usepackage{inconsolata}

\usepackage{graphicx}
\usepackage{array}
\usepackage{booktabs}

\newcommand{\hflogo}{\raisebox{-0.25\height}{\includegraphics[height=1.15em]{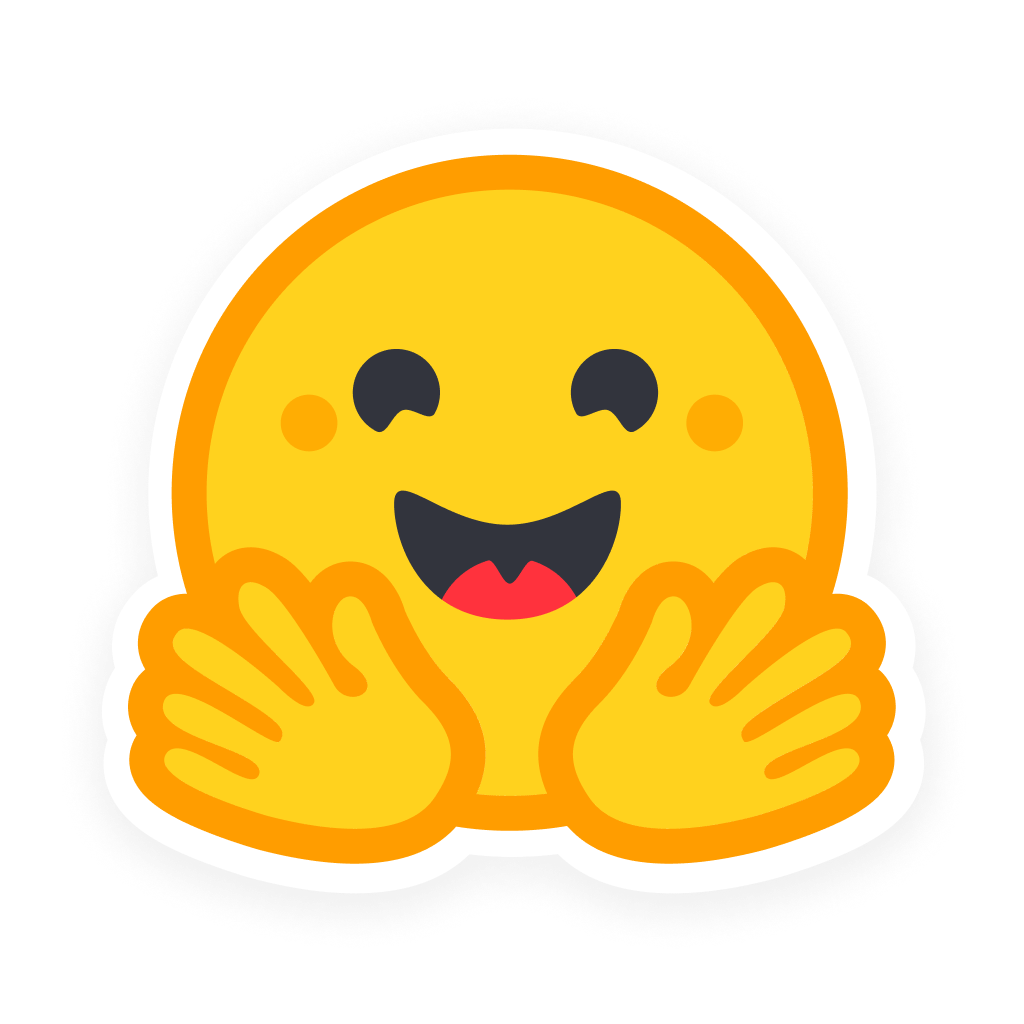}}}
\newcommand{\hfdataset}{%
  \href{https://huggingface.co/datasets/MU-NLPC/Edustories-en}%
       {\hflogo\hspace{0.35em}\texttt{MU-NLPC/Edustories-en}}}

\usepackage{twemojis}

\newcommand{\weblogo}{\raisebox{-0.25\height}{\includegraphics[height=1.15em]{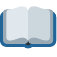}}}
\newcommand{\website}{%
  \href{https://edustories.cz/en}%
       {\weblogo\hspace{0.35em}\texttt{https://edustories.cz}}}
\title{Edustories: A Collection of Real-world Case Studies \\from Classroom Practices}

\author{
 \textbf{Michal Štefánik\textsuperscript{1,2}}\ \ \ 
 \textbf{Jan Nehyba\textsuperscript{1}}\ \ \ 
 \textbf{Jirina Karasova\textsuperscript{1}}\ \ \ 
 \textbf{Martin Fico\textsuperscript{1}}\ \ \ 
\\
 \textbf{Lucie Škarková\textsuperscript{1}}\ \ \ 
 \textbf{Markéta Košatková\textsuperscript{1}}\ \ \ 
 \textbf{David Kosatka\textsuperscript{1}}
\\
\\
 \textsuperscript{1}Faculty of Education, Masaryk University, Czechia \\
 \textsuperscript{2}R\&D Center for Large Language Models, National Institute of Informatics, Japan \vspace{10pt}
\\
   \hfdataset\ \ \ \ \ \ \ \website
}

\begin{document}
\maketitle

\begin{abstract}

Despite the widely recognized potential of AI in education, most prior work has focused on individualized student assistance. In contrast, the majority of educational practice worldwide still takes place in collective classroom settings. To enable researchers to study AI assistance in collective teaching, we introduce Edustories, a dataset of \textbf{1,492 teacher-written case studies} describing real elementary and high-school classroom situations involving challenging student behavior, pedagogical interventions, and their outcomes. 
Among many other applications, Edustories enables evaluating LLMs' ability to predict the success of teacher interventions, crucial for providing practicing teachers with useful feedback. Comparing the latest models from four language-model families against expert assessments, we find that current models fall short of human expertise in predicting classroom outcomes; the strongest models reach 58\% accuracy compared to 64\% of human experts. This gap highlights both the limitations and the emerging potential of AI as assistants for practicing teachers.

\end{abstract}

\section{Introduction}

Artificial intelligence has a large potential to support and improve educational practice with over one third of teachers already using AI in their daily practice to prepare materials or personalize exercises \cite{WaltonAI2024}. Prior work has demonstrated that AI-driven systems can accelerate learning and improve outcomes in a variety of educational domains, including intelligent tutoring systems, automated feedback for programming and mathematics, and adaptive learning platforms \cite{MATOS2025100571}. However, much of this progress has focused on personalized tutoring, where models interact with a single learner and generate personalized explanations, hints, or corrections \cite{WANG2024124167}.

\begin{figure}
    \centering
    \!\!\!\!\!\!\!\!\!\includegraphics[width=1.15\linewidth]{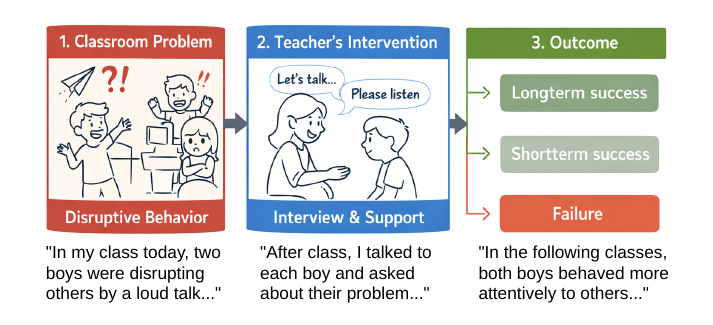}\vspace{-3pt}
    \caption{Edustories is a dataset of authentic classroom situations, teachers' interventions and their outcomes. \vspace{-10pt}}
    \label{fig:placeholder}
\end{figure}

In contrast, collective classroom instruction—in which a teacher simultaneously engages with many students—remains the dominant educational paradigm in elementary and secondary schools. Practicing teachers in such settings face challenges that go beyond individual learning trajectories, including classroom management, social dynamics among students, and the need to balance instructional goals with behavioral interventions \cite{Kearney2019TheCO}. Despite its central role in everyday education, relatively little prior work has aimed to develop automated systems that explicitly support teachers in managing and reflecting on collective lessons \cite{WANG2024124167}.

We argue that this gap can be attributed to two main factors. First, collective lessons are inherently more complex than one-on-one educational interactions, as they involve multiple actors, social relationships, and context-dependent dynamics that are difficult to model computationally \cite{Toom2021}. Second, the collection of data documenting real classroom situations poses significant challenges: strict ethical and legal requirements for protecting the anonymity of students and practicing teachers limit the availability of detailed, real-world descriptions without compromising authenticity or completeness \cite{educsci15040500}. These constraints likely explain why, to best of our knowledge, there is no other public dataset documenting situations from collective classroom instruction.

To address this gap and to support the development of AI technologies for assisting teachers in collective lessons, we introduce Edustories, a dataset of \textbf{1,492 teacher-authored descriptions} of challenging situations encountered during classroom instruction. Each case study includes a narrative description of the situation, the teacher’s chosen response or intervention, and the implied outcome, reflecting the teacher’s own assessment to what extent the intervention was successful.

Edustories enables a wide range of applications beyond educational AI, including \textbf{cognitive science} (e.g., analyzing student diagnoses and hobbies) and \textbf{developmental psychology} (e.g., studying how behavior evolves with age). In this work, we focus on evaluating a core ability of AI necessary in providing valuable situational feedback to teachers: \textbf{distinguishing between successful and unsuccessful strategies in classroom interventions}. We find that modeling outcomes of classroom situations remains challenging even for the strongest LLMs, especially when distinguishing short- vs. long-term effects, with 5 of 6 models performing below the lower bound of expert assessment.

\subsection{Related collections}

\noindent\textbf{Classroom behaviour\ \ }
Several datasets study classroom behaviour using video recordings of teaching practice, including MM-TBA \cite{Huang2025AMD}, TBU \cite{CAI2025104376}, and FSCB \cite{app14146153}, with the latter two also providing explicit behaviour labels. These datasets capture what happens in the classroom at an observable level, but they are limited to video data. As a result, they do not represent deeper situational factors captured in Edustories -- such as students’ prior history, earlier interactions, or how teachers interpretation -- essential for pedagogical decision-making and providing a meaningful feedback.

\noindent\textbf{Teacher reflections\ \ }
Teacher reflection has been explored through textual datasets aimed at analysing affective and cognitive aspects of teaching practice. The CEReD dataset \cite{Stefanik2021CEReD} consists of teacher reflection texts annotated for emotional and reflective elements and has been used for emotion classification and reflective analysis \cite{Nehyba2023applications,DBLP:journals/eait/ZhangHPG24}, often with the goal of providing feedback to teachers on their practice \cite{educsci15101315,zhang24_phdthesis}. However, existing datasets are designed for classification and do not preserve complete classroom narratives. In contrast, Edustories provides holistic case descriptions that retain the rich semantics and narrative of class situations.

\noindent\textbf{Narrative datasets} that model the evolution of situations and the consequences of actors’ actions have been developed in several non-educational domains. Examples include datasets focused on moral decision-making and plausibility reasoning \cite{Emelin2021MoralSS,bae-etal-2025-charmoral}, historical narratives \cite{Leteno2025HistoiresMA}, social interactions and norms \cite{forbes-etal-2020-social}, and ethical judgments \cite{hendrycks2021ethics}. While these resources share a narrative structure similar to Edustories, none of them focus on educational practice.

\section{The Edustories Dataset}
\vspace{-4pt}


\begin{table*}[t]
\centering
\vspace{-12pt}
\resizebox{0.75\textwidth}{!}{%
\begin{tabular}{lccccc}
\hline
Feature & Num. filled & Avg. words & Avg. sentences & Num. categories & Most common \\
\hline
Description      & 1492 & 141.39 & 9.41 & -- & -- \\
Anamnesis        & 1491 & 79.55  & 5.86 & -- & -- \\
\quad category   & 1478 & --     & --   & 14 & verbal disruption (12.24\%) \\
Intervention     & 1492 & 134.99 & 8.21 & -- & -- \\
\quad category   & 1475 & --     & --   & 15 & interview (30.35\%) \\
Outcome          & 1492 & 82.78  & 5.26 & -- & -- \\
\quad category   & 1475 & --     & --   & 3  & long-term success (44.63\%) \\
Student age      & 1455 & --     & --   & -- & 12 years \\
Hobbies          & 1365 & 4.75   & --   & -- & computer games (8.17\%) \\
Diagnosis        & 583  & --     & --   & 72 & ADHD (22.52\%) \\
Disorders        & 908  & --     & --   & 103& lying (24.66\%) \\
\hline
\end{tabular}%
}
\caption{Overview of statistics of the main features of Edustories data collection (1,492 samples in total).\vspace{-12pt}}

\end{table*}

\paragraph{Origin and Preparation}
The case studies contained in the Edustories dataset were written by 241 trainee teachers during their teaching practice at Czech elementary and high schools, coordinated between 2023 and 2026. These trainee teachers are finishing university students completing classroom practice as the last part of their curricula, collecting their case studies with the primary goal of receiving feedback in their collective reflection sessions that follow their teaching practice.

The trainee teachers collect their stories in a structured format designed to capture key aspects of classroom situations. Mainly, the collected stories include (1)~a narrative description of the situation, (2)~an anamnesis providing background information about the student playing a central role in the story, (3)~the problem the teacher had to address, (4)~the intervention applied by the teacher as a response to the situation, and (5)~a description of the perceived outcome of the intervention. Together with textual descriptions, authors of the stories also report the age of the student, their hobbies, and the presence of any reported disorders or diagnoses.


Based on the original textual descriptions, the collected stories were then enriched with categorical features labelled and peer-reviewed by two independent education experts, where each of the expert annotated a half of the cases and peer-reviewed the other half, with all disputable cases being resolved through expert discussion. Such annotated features categorise the type of problem (Anamnesis), the action taken in the story (Intervention) and the observed effect (Outcome).

To make the dataset accessible to the broader AI and NLP research community, the free-text components of Edustories (description, anamnesis, problems, and interventions) were additionally translated from Czech to English. The translation was performed using the service chosen in the expert-based evaluation detailed in Section~\ref{appx:translation}.

\paragraph{Consent and Anonymization}
To protect the privacy of all participants, the free-text case descriptions underwent a two-stage anonymization process. First, all personal names and explicit (non-relative) references to dates, times, and locations were automatically identified and randomized using the UFAL Named Entity Recognition service tailored for Czech \cite{SevcikovaEtAl2007CNEC}. Second, the anonymized texts were manually reviewed during peer-reviewed quality control and augmentation to ensure that no identifying information remains.
All authors of Edustories case studies voluntarily provided written informed consent
permitting the further use of their data in anonymized form.\footnote{The proposal of data collection first underwent review by the ethical committee of the coordinating organisation (to be specified in a non-anonymous version) with a positive result.}

\section{Experiments: Predicting Success}
One of the most direct and impactful applications of AI in improving collective classroom instruction is the provision of feedback to teachers' instructional and intervention strategies. Crucially, this application depends on AI's core ability to distinguish successful and unsuccessful strategies. 

Edustories provides complete descriptions of classroom situations, teachers' proposed interventions and categorical annotations of intervention success. This enables us to benchmark the ability of LLMs to model the intervention outcome, necessary for providing teachers with a feedback that could improve the likelihood of their success. In this study, we use these features to assess the performance of six state-of-the-art language models from five model families, and compare their accuracy to evaluations provided by educational experts.

\subsection{Experimental Setup}
\label{sec:experimental_setup}
Our setup aims to align as closely as possible with the likely real-world usage of LLMs by lay practicing teachers, while giving experts the freedom to approach the problem with their best knowledge.

\paragraph{Language models}
For each case study, we construct evaluation prompts by incorporating the \textit{Description}, \textit{Anamnesis}, and \textit{Intervention} fields into the task description. We experiment with five alternative prompt formulations and select the one that performs best across models, noting that differences between variants are relatively small and fall within the evaluation confidence interval (±1.6\%). Thus, to favor transparency and reproducibility, we use a matching prompt for all models evaluations (detailed in Appendix~\ref{appx:eval_prompt}).

We evaluate a set of open-source language models that represent the state of the art across a range of sizes and model families. As a proxy for model quality, we select models that lead their respective size categories on the MMLU-Pro benchmark \cite{hendrycks2021measuring,wang2024mmlupro}. The evaluated models span the most recent models from Llama 3, Qwen 3, Mistral v0.3 and Olmo 3 families (see Appendix~\ref{appx:models} for model IDs).


\paragraph{Expert judgements}


We compare outcome prediction ability of LLMs with five independent educational experts on 310 case studies from Edustories representatively covering the range of students' age. Upon a mutual discussion, the experts choose to tackle this problem by constructing their final assessment from individual judgments of three facets of Behavior Management Frameworks\footnote{Detailed in Appendix~\ref{appx:behavior-management-frameworks}.} \cite{Cipriano2023,BaturaiteBunka2024,Bradshaw2012,Fronius2019}: (1) \textbf{Humanistic orientation}, reflecting emphasis on empathy, dialogue, and student needs versus reinforced control; (2) \textbf{Reactivity}, capturing \textit{reactive} (cf. \textit{preventive}) strategies; and (3) \textbf{Systematicity}, distinguishing generalizable strategies with contextual adaptation from incident-specific actions. Experts rate each of these facets on a centered Likert scale (a score between -2 and +2) common in Education studies (\citet{Hudon2025ClinicalReasoningSCT,Pilitti2023HigherEd}; inter alia) and finally construct the overall \textbf{Appropriateness} of each intervention on the same scale.
These judgements are then mapped to the Edustories outcome labels by maintaining their zero-centered semantics: -2 (i.e. decisively negative) corresponding to \textit{failure}, -1 and 0 (partially negative but non-positive) corresponding to \textit{partial or short-term success} and +1 and +2 (positive or perfect) corresponding to \textit{long-term success}.
Expert labels agree in 82,56\% of cases, with average inter-annotator $\kappa=0.7356$, evidencing high agreement.


\subsection{Results}

\begin{figure}
    \centering
    \scalebox{0.9}{
        \includegraphics[width=\linewidth]{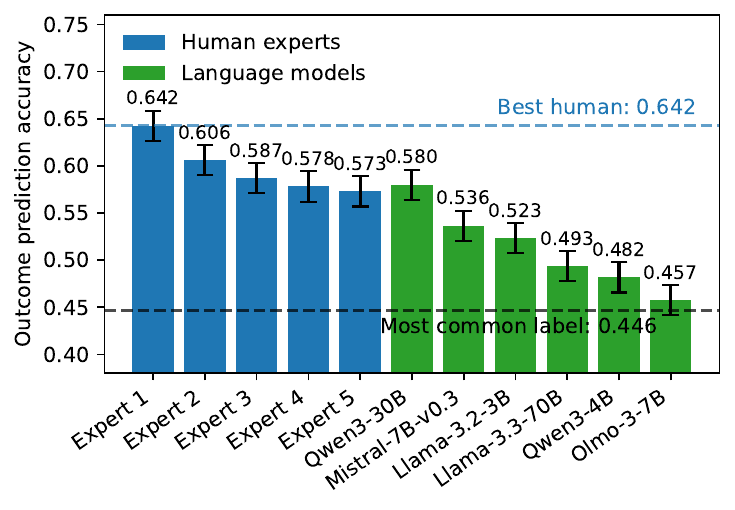}
        
    }
    \vspace{-9pt}
    \caption{Accuracy of outcome prediction for interventions and case studies of Edustories dataset, for (blue) appropriateness evaluation of human experts and (green) direct prediction of selected language models. Evaluation on n=310 samples to allow comparability with judgements of human experts (§\ref{sec:experimental_setup}). Confidence interval estimates for $\alpha=0.1$ covering ±1.6\% ranges.\vspace{-10pt}}
    \label{fig:main_results}
\end{figure}

Figure~\ref{fig:main_results} compares outcome prediction accuracy between human experts and selected language models. Among all evaluated models, only one model (Qwen-3-30B) reaches performance overlapping with human expert accuracy, achieving 0.580 compared to 0.573–0.587 for the lower-performing experts. Qwen-3-30B also exhibits a clear margin over all other language models.

Our results indicate that model size is not the primary performance factor; For example, the larger Llama-3.3-70B underperforms the smaller Llama-3.2-3B, suggesting that architectural choices, training data and other training refinements play a more important role than scale alone. This has positive implications for practical deployment. We hypothesise that outcome prediction may depend more on general reasoning capability, as the two best-performing models also show particularly strong performance on reasoning-focused tasks \cite{yang2025qwen3technicalreport,toh2025not,morozov-etal-2025-fast}.

\noindent \textbf{Comparative analysis\ \ }
A closer comparison of the best-performing model and human expert (Appendix~\ref{appx:confusion_matrices}) reveals qualitatively different error patterns. Human experts’ errors predominantly arise from assigning failure cases to short-term success, while they remain highly accurate (84.41\%) in identifying long-term success. In contrast, Qwen-3-30B and other language models primarily struggle to distinguish between short- and long-term success.

This pattern is consistent across experts, who substantially outperform language models in identifying long-term success (71.97\% vs.\ 49.88\%), though they are more susceptible to other types of errors. Overall, performance variance is considerably higher among language models (0.457–0.580) than among human experts (0.573–0.642).

Taken together, these results indicate that human experts remain more reliable and consistent, despite the near-expert performance achieved by the best-performing language model. Nevertheless, the complementary nature of errors made by humans and models suggests that current language models may already provide value as assistive tools rather than replacements for educational experts.

\vspace{-3pt}
\section{Conclusions}
\vspace{-3pt}

We introduce Edustories, a new collection of 1,492 authentic, teacher-written case studies from elementary and high-school classrooms. The dataset was proofread and anonymized through a two-stage process that included human review, making it the first available educational dataset documenting classroom practices at this level of detail.

Edustories enables a wide range of research directions in education, linguistics, and AI, including sentiment analyses, identification of predictive features of successful interventions, and evaluation of AI assistants in collective educational practice. In this work, we focus on one such direction: assessing the ability of current LLMs in providing feedback on teachers’ proposed interventions.

Our results show that while the best-performing models can reach the lower bound of human expert performance, language models remain less consistent and exhibit different failure modes than humans. These findings highlight the importance of specialized educational benchmarks before using AI assistants in practice, but also underscore their existing potential for educational practice.

\section*{Limitations}

We acknowledge several limitations of our work that point to directions for future research. First, our evaluation of outcome prediction does not fully capture the complexity of real-world deployment of assistive AI in educational settings. While we focus on outcome prediction as a key indicator of practical viability, effective integration into practice would likely require a more sophisticated pipeline, supporting teacher-facing feedback, such as free-form rationales or conversational interfaces. Although this lies beyond the scope of our current contribution, we view it as an impactful and interesting direction for future work at the intersection of AI and human–computer interaction.

Second, we note the narrow demographic group of teachers participating in our data collection. All stories originate from the practicing teachers in the last years of their university study. As such, our dataset may not fully capture the strategies and solutions of older generations of teachers.
Further, we acknowledge the limitation of Edustories dataset in the quality of English translation of free-form texts, constrained by the precision of the commercial translation services.
While we pick the translation service for free-form texts systematically based on human assessment from among Google Translate, DeepL and Open-source Czech-English translation models, we note that errors may occus and estimate that 5--10\% of stories may contain a type or ``translationese'' text.
As we do not dispose of sufficient power of human proofreaders to manually verify all the English versions of our stories, we will appreciate the effort of the community in helping us to identify cases where the quality of free-form features of Edustories could be improved and possibly helping us to address them.

Finally, we acknowledge the limitation of the scope our evaluated models, that do not include the models over 100-billion parameters due to our computational constraints.
Note that we intentionally exclude the proprietary API AI services from evaluations as these may not present an apples-to-apples comparison to raw language models evaluated locally.
Furthermore, we believe that controlled, locally-deployable models correspond better with the practical application of feedback delivery systems as these systems will operate directly with sensitive data from classrooms that may not be safely shared with third parties.


\bibliography{custom}

\appendix

\section{Experimental details}
\label{sec:appendix}

\subsection{Confusion matrices}
\label{appx:confusion_matrices}

\begin{table*}[tbh]
\centering

\begin{minipage}[t]{0.48\textwidth}
\centering
\scalebox{0.85}{
\setlength{\tabcolsep}{6pt}
\renewcommand{\arraystretch}{1.2}
\begin{tabular}{llccc}
\toprule
 & & \multicolumn{3}{c}{\textbf{Predicted outcome}} \\
\cmidrule(lr){3-5}
\textbf{True outcome} &  & \textbf{fail} & \textbf{success-short} & \textbf{success-long} \\
\midrule
\textbf{fail}          &  & 0.00  & 91.67 & 8.33  \\
\textbf{success-short} &  & 7.69  & 27.88 & 64.42 \\
\textbf{success-long}  &  & 2.69  & 12.90 & 84.41 \\
\bottomrule
\end{tabular}
}
\caption{\textbf{Best expert:} Row-normalized confusion matrix (\%) for three-way outcome classification
for the best-performing human expert.}
\label{tab:confmat_best_human}
\end{minipage}
\hfill
\begin{minipage}[t]{0.48\textwidth}
\centering
\scalebox{0.85}{
\setlength{\tabcolsep}{6pt}
\renewcommand{\arraystretch}{1.2}
\begin{tabular}{llccc}
\toprule
 & & \multicolumn{3}{c}{\textbf{Predicted outcome}} \\
\cmidrule(lr){3-5}
\textbf{True outcome} &  & \textbf{fail} & \textbf{success-short} & \textbf{success-long} \\
\midrule
\textbf{fail}          &  & 41.67 & 8.33  & 50.00 \\
\textbf{success-short} &  & 12.62 & 41.75 & 45.63 \\
\textbf{success-long}  &  & 7.57  & 24.32 & 68.11 \\
\bottomrule
\end{tabular}
}
\caption{\textbf{Best language model:} Row-normalized confusion matrix (\%) for three-way outcome classification
for the best-performing \textbf{language model} (Qwen3-30B).}
\label{tab:confmat_best_lm}
\end{minipage}

\end{table*}

\begin{table*}[tbh]
\centering

\begin{minipage}[t]{0.48\textwidth}
\centering
\scalebox{0.85}{
\setlength{\tabcolsep}{6pt}
\renewcommand{\arraystretch}{1.2}
\begin{tabular}{llccc}
\toprule
 & & \multicolumn{3}{c}{\textbf{Predicted outcome}} \\
\cmidrule(lr){3-5}
\textbf{True outcome} &  & \textbf{fail} & \textbf{success-short} & \textbf{success-long} \\
\midrule
fail          &  & 31.67 & 55.00 & 13.33 \\
success-short &  & 17.49 & 28.95 & 53.56 \\
success-long  &  & 5.87  & 22.16 & 71.97 \\
\bottomrule
\end{tabular}
}
\caption{\textbf{All experts:} Row-normalized confusion matrix (\%) averaged across all human experts
for three-way outcome classification.}
\label{tab:confmat_avg_humans}
\end{minipage}
\hfill
\begin{minipage}[t]{0.48\textwidth}
\centering
\scalebox{0.85}{
\setlength{\tabcolsep}{6pt}
\renewcommand{\arraystretch}{1.2}
\begin{tabular}{llccc}
\toprule
 & & \multicolumn{3}{c}{\textbf{Predicted outcome}} \\
\cmidrule(lr){3-5}
\textbf{True outcome} &  & \textbf{fail} & \textbf{success-short} & \textbf{success-long} \\
\midrule
fail          &  & 45.83 & 45.83 & 8.33  \\
success-short &  & 14.06 & 54.51 & 31.44 \\
success-long  &  & 8.29  & 41.83 & 49.88 \\
\bottomrule
\end{tabular}
}
\caption{\textbf{All language models:} Row-normalized confusion matrix (\%) averaged across all language models
for three-way outcome classification.}
\label{tab:confmat_avg_lms}
\end{minipage}

\end{table*}

Tables~\ref{tab:confmat_best_human} and \ref{tab:confmat_best_lm} shows classification confusion matrix for the best-performing human annotator and best-performing language model, respectively. Tables~\ref{tab:confmat_avg_humans} and \ref{tab:confmat_avg_lms} shows classification confusion matrix aggregated over all human experts, and evaluated language models, respectively.
Note that the proportions of labels in our evaluation set are as follows: Long-term success: 0.446, Short-term success: 0.244, Failure: 0.310.

\subsection{Evaluation Prompt}
\label{appx:eval_prompt}

Table~\ref{appx_table:eval_prompt} displays the exact form of the best-performing prompt we applied in our evaluations. The selection process is detailed in Section~\ref{sec:experimental_setup}.

\begin{table}[h!]
\resizebox{0.47\textwidth}{!}{%
\begin{tabular}{l}
\hline
\textbf{Prompt}  \\
\begin{tabular}[c]{@{}l@{}}You will be given a short case study written by a teacher. \\ It describes a situation and the intervention chosen.\\ \\ Your task is to classify the expected overall outcome of \\the solution into exactly one category:\\ 'Longterm success', 'Partial success', 'failure' or 'I don't know'\\ \\  Important rules:\\ - Choose exactly one category.\\ - Reply with the category name ONLY, copied exactly.\\ - No punctuation, explanation, or extra text.\\ \\ Case study:\\ \\ """\{Description\} \{Anamnesis\} \{Intervention\}"""\\ \\
\end{tabular}  \\
\textbf{Response}  \\
One of \{'Longterm success' / 'Partial success'/ 'failure' / 'I don't know'\} \\
\hline
\end{tabular}%
}

\caption{Final prompt we use in our evaluations. Chosen as the best-performing variant among 5 candidates.}
\label{appx_table:eval_prompt}
\end{table}

\subsection{Evaluated models}
\label{appx:models}

In our experiments, we specifically evaluate the following models with HuggingFace identifiers: Qwen3-30B and Qwen3-4B introduced by \citet{yang2025qwen3technicalreport} corresponding to \textit{Qwen3-30B-A3B-Instruct-2507} and \textit{Qwen3-4B-Instruct-2507}, Llama 3.2-70B and Llama 3.3-3B introduced by \cite{grattafiori2024llama3herdmodels} corresponding to \textit{meta-Llama-3.3-70B-Instruct} and \textit{Llama-3.2-3B-Instruct}, Mistral-7B introduced by \citet{jiang2023mistral7b} corresponding to \textit{Mistral-7B-Instruct-v0.3} and Olmo-3 introduced by \citet{olmo2025olmo3} corresponding to \textit{Olmo-3-7B-Instruct}.

\subsection{Translation process}
\label{appx:translation}

To make Edustories accessible to a wider scientific community, we additionally translated the anonymised free-text features from Czech to English. We experimented with four translation tools available to date\footnote{As of November, 2025}: (1) Open-source Opus translation models \cite{tiedemann-thottingal-2020-opus}, Google Translate API\footnote{\url{https://translate.google.com}}, DeepL\footnote{\url{https://deepl.com/en/translator}} and GPT 5.2 Thinking\footnote{\url{https://chatgpt.com/}}. For a randomly-chosen set of 50 samples from our datasets, we manually verified the fluency and semantic equivalence to the source texts. Both of these cases were minor (i.e. not breaking comprehensibility) and occurred in the Intervention texts. We found that the GPT 5.2 API performs on par with DeepL in terms of semantic equivalence -- in 50 samples, annotators identified 2 cases where the semantic equivalence was not perfect but with minor mistakes. However, we found that the GPT 5.2 outperforms other services in fluency, where our annotators did not identify any mistakes among the 50 samples. Based on this evaluation, all translations provided in Edustories were produced by GPT 5.2. Note that the original features (in Czech) are retained in the collection and can be used for further improving the translation quality with future translation models or as the original source of translation to other languages.

\subsection{Overview of Selected Behavior Management Frameworks}
\label{appx:behavior-management-frameworks}

This section summarizes the behavior management frameworks that the educational experts involved in our study identified as most suitable for their structured decomposition of classroom behavior into three evaluation dimensions and to their final ratings.

\paragraph{Social and Emotional Learning (SEL).}
SEL aims to develop students' self-awareness, self-management, social awareness, relationship skills, and responsible decision-making. Large-scale meta-analyses indicate that universal, school-based SEL programs improve students' social--emotional skills, behavior, and academic achievement at posttest \citep{durlak2011impact}, and that these benefits can persist months to years after the intervention \citep{taylor2017promoting}. A more recent meta-analysis further reports positive effects on well-being, conduct problems, and school functioning, while also highlighting implementation quality as an important moderator of outcomes \citep{Cipriano2023}.

\paragraph{Nonviolent Communication (NVC).}
NVC is a structured approach to communication organized around observations, feelings, needs, and requests \citep{rosenberg2015nonviolent}. In classroom settings, it is used to de-escalate conflict and model empathic, need-focused dialogue, including through teacher ``I-messages,'' perspective-taking, and collaborative requests. Although the causal evidence base in K--12 classrooms remains limited, quasi-experimental and education- or youth-training studies suggest that NVC-based programs can improve empathy and communication-related competencies \citep{BaturaiteBunka2024,sung2022effects}.

\paragraph{Positive Behavioral Interventions and Supports (PBIS).}
PBIS is a school-wide, tiered prevention framework in which expected behaviors are explicitly defined, taught, acknowledged, and monitored using data-informed supports across Tiers 1--3. In a cluster-randomized effectiveness trial across 37 elementary schools, PBIS reduced behavior problems and office discipline referrals and improved prosocial functioning, with the strongest effects observed among students whose exposure began in kindergarten \citep{Bradshaw2012}.

\paragraph{Restorative Practices (RP).}
Restorative practices frame misbehavior as harm to people and relationships, emphasizing both accountability and repair through approaches such as circles, conferences, and re-integration plans. A district-wide randomized trial in Pittsburgh found that RP implementation reduced suspensions, particularly in elementary grades, although effects on achievement and school climate varied by outcome and grade band \citep{augustine2018can}. Systematic reviews describe school-based RP as promising but implementation-sensitive, and emphasize the need for additional rigorous causal evidence \citep{Fronius2019}.

\section{Use of AI Assistants}
\label{appx:ai_assistants}

We acknowledge the use of AI assistants during the creation of this paper, namely in plots refinements, tables formatting, grammar control and text polishing. AI assistants were not employed in the creation of ideas, results and connections described in this paper.

\end{document}